\documentclass[lettersize,journal]{IEEEtran}

\usepackage{graphicx}
\usepackage{amsmath}
\usepackage{amssymb}
\usepackage{booktabs}
\usepackage{caption}
\usepackage{algorithm}
\usepackage{algorithmic}
\usepackage{xcolor}
\usepackage{diagbox}
\usepackage[export]{adjustbox}
\usepackage[pagebackref,breaklinks,colorlinks]{hyperref}

\begin{document}

\title{Is Bio-Inspired Learning Better than Backprop? \textit{\\Benchmarking Bio Learning vs. Backprop}}

\author{

Manas Gupta\textsuperscript{1}, Sarthak Ketanbhai Modi\textsuperscript{2}, Hang Zhang\textsuperscript{3}, Joon Hei Lee\textsuperscript{4}, Joo Hwee Lim\textsuperscript{1,4,5}

\vspace{3mm}

~\IEEEmembership{
\small \textsuperscript{1}Institute for Infocomm Research (I2R), A*STAR, Singapore, \\
\textsuperscript{2} School of Mechanical and Aerospace Engineering, Nanyang Technological University (NTU), Singapore, \\
\textsuperscript{3} Faculty of Science, National University of Singapore (NUS), Singapore, \\                 \textsuperscript{4} School of Computer Science and Engineering, Nanyang Technological University (NTU), Singapore, \\
\textsuperscript{5} Centre for Frontier AI Research (CFAR), A*STAR, Singapore\\}

\vspace{3mm}

\small manas\_gupta@i2r.a-star.edu.sg, sarthak005@e.ntu.edu.sg, zhanghang@u.nus.edu, jlee263@e.ntu.edu.sg, joohwee@i2r.a-star.edu.sg

}

\maketitle

\begin{abstract}
Bio-inspired learning has been gaining popularity recently given that Backpropagation (BP) is not considered biologically plausible. Many algorithms have been proposed in the literature which are all more biologically plausible than BP. However, apart from overcoming the biological implausibility of BP, a strong motivation for using Bio-inspired algorithms remains lacking. In this study, we undertake a holistic comparison of BP vs. multiple Bio-inspired algorithms to answer the question of whether Bio-learning offers additional benefits over BP. We test Bio-algorithms under different design choices such as access to only partial training data, resource constraints in terms of the number of training epochs, sparsification of the neural network parameters and addition of noise to input samples. Through these experiments, we notably find two key advantages of Bio-algorithms over BP. Firstly, Bio-algorithms perform much better than BP when the entire training dataset is not supplied. Four of the five Bio-algorithms tested outperform BP by upto 5\% accuracy when only 20\% of the training dataset is available. Secondly, even when the full dataset is available, Bio-algorithms learn much quicker and converge to a stable accuracy in far lesser training epochs than BP. Hebbian learning, specifically, is able to learn in just 5 epochs compared to around 100 epochs required by BP. These insights present practical reasons for utilising Bio-learning beyond just their biological plausibility and also point towards interesting new directions for future work on Bio-learning.

\end{abstract}

\section{Introduction}
\label{intro}

\begin{figure*}
\centering
\includegraphics[width=11cm]{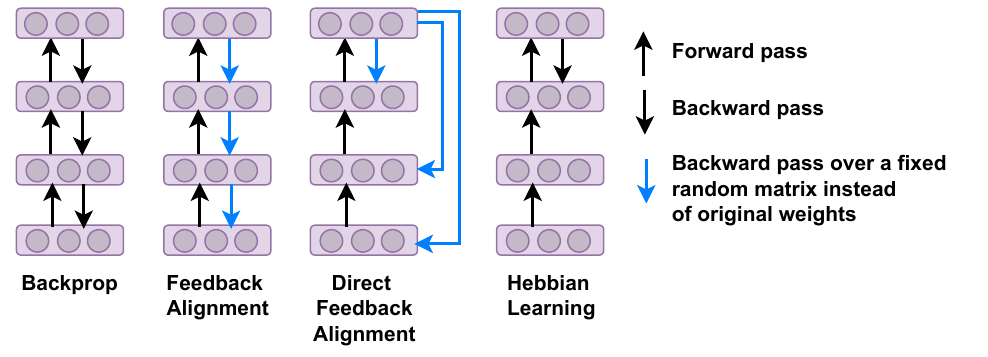}
\caption{Illustration of how different learning algorithms. Feedback Alignment and Direct Feedback Alignment back-propagate gradients over a fixed random matrix instead of the original weights, thereby relaxing the condition of having the same pathway (weights) for both the forward and backward pass. Hebbian learning does not need any gradients to be back-propagated as it learns in an unsupervised manner. Only the last layer, which is the classification layer, learns based on gradients.}
\label{fig:algos}
\end{figure*}
x
Back-propagation (BP) \cite{Rumelhart1986LearningRB} has been the mainstay of training deep neural networks over the past decade. However, it is not considered biologically plausible \cite{10.1145/3354265.3354275, Lillicrap2020BackpropagationAT}. Hence, research interest has always remained in finding learning algorithms that the human brain might be using. Over the last few years a myriad of Bio-inspired algorithms have been proposed. These algorithms have now achieved comparable accuracy as Backprop (BP) and also have been known to be applied to a vast array of deep learning tasks such as computer vision, reinforcement learning, language modelling, amongst others. 

Bio-algorithms differ from BP in a few crucial aspects. Firstly, weight symmetry is required by BP between the forward and backward connections, that is, the same weight pathways must be used for both the forward and backward passes. This is not biologically plausible in the brain due to the uni-directional nature of brain synapses and is called the weight transport problem \cite{GROSSBERG198723}. Secondly, the problem of update locking occurs in BP. That is all layers must complete their processing before the backward pass can begin and the weights can be updated. This does not appear to be biological in nature as many plasticity algorithms in the brain learn in a local fashion, that is, learn between two pairs of neurons itself, without depending upon the activity of other layers \cite{BALDI2017110}. Thus, the brain supports multiple sets of neurons to learn in parallel. This also brings us to the third crucial difference, which is that BP learns in a global fashion, by calculating an error at the end of the last layer and then feeding it back to each layer. On the contrary, the brain learns in a local fashion, dependent only upon the activity of two connected neurons without a global loss to be learnt and backpropagated \cite{MAASS19971659}.  

The motivation of using and developing Bio-algorithms mostly remains to make these algorithms more biologically plausible than BP. Tangible benefits over BP have not yet been demonstrated and therefore, adoption of Bio-learning algorithms outside the Bio-learning community remains small. We take a step in this direction and conduct a holistic benchmarking of Bio algorithms vs. BP in various experimental settings, to understand the practical benefits of Bio-algorithms. We look at popular Bio-learning algorithms like Feedback Alignment (FA) \cite{Lillicrap2016RandomSF}, Direct Feedback Alignment (DFA) \cite{DFA}, Hebbian Learning (HB) \cite{Krotov7723, Amato2019HebbianLM, 9207242, hebbnet, miconi}, Predictive Coding (PC) \cite{PC} and Difference Target Propagation (DTP) \cite{TargetProp}. Many other Bio-algorithms exist but we choose these as they have gained traction over the years and serve as representative samples from the different categories of Bio-algorithms.  

For our benchmarking scenarios, we include a range of aspects important for practical usage of a learning algorithm. We measure performance of the algorithms under i) Limited availability of training data, ii) Limited training budget (in terms of the number of training epochs), iii) Application of noise to input samples, and iv) Application of sparsity to the parameters of the base neural network models (i.e. pruning the parameters of the model). In this paper, we have the following key contributions-

\begin{enumerate}
    \item We show that Bio-algorithms perform surprisingly well in the scenario of limited training data. Four of the five Bio-algorithms tested, namely, FA, DFA, HB and PC outperform BP when only 20\% of data is available for testing on multiple datasets. 
    \item Another interesting finding is that some Bio-inspired algorithms perform exceedingly well in scenarios of limited training budgets. Specifically, HB and DTP learn much faster than BP and arrive at high stable accuracies in far lesser training epochs than BP. HB especially, learns within 5 epochs on multiple datasets where BP takes around 100 epochs to learn. 
    \item To understand this superior performance even further, especially for the case of Hebbian learning, we plot the features learnt by BP vs. HB. We find that HB learns much more fine-grained and explainable features whereas BP seems to learn features that are much more random looking. We postulate that this might be one of the underlying reasons for the better learning capability of HB. 
    \item We also find that Bio-learning algorithms are robust against different kinds of noises and also adept under the application of parameter sparsity to the neural networks, achieving similar accuracy as BP.
\end{enumerate}

The above findings, especially points 1 and 2, are important advantages of Bio-learning algorithms that make them very useful in learning with less data applications or resource-constrained training environments like real-time devices. They also motivate the need to work on Bio-learning algorithms for practical advantages and beyond the singular motivation of solving the biological implausibility of Backprop.

\section{Related Work}

\begin{figure*}[h]
\centering
\begin{minipage}{7cm}
\centering
\includegraphics[width=5cm]{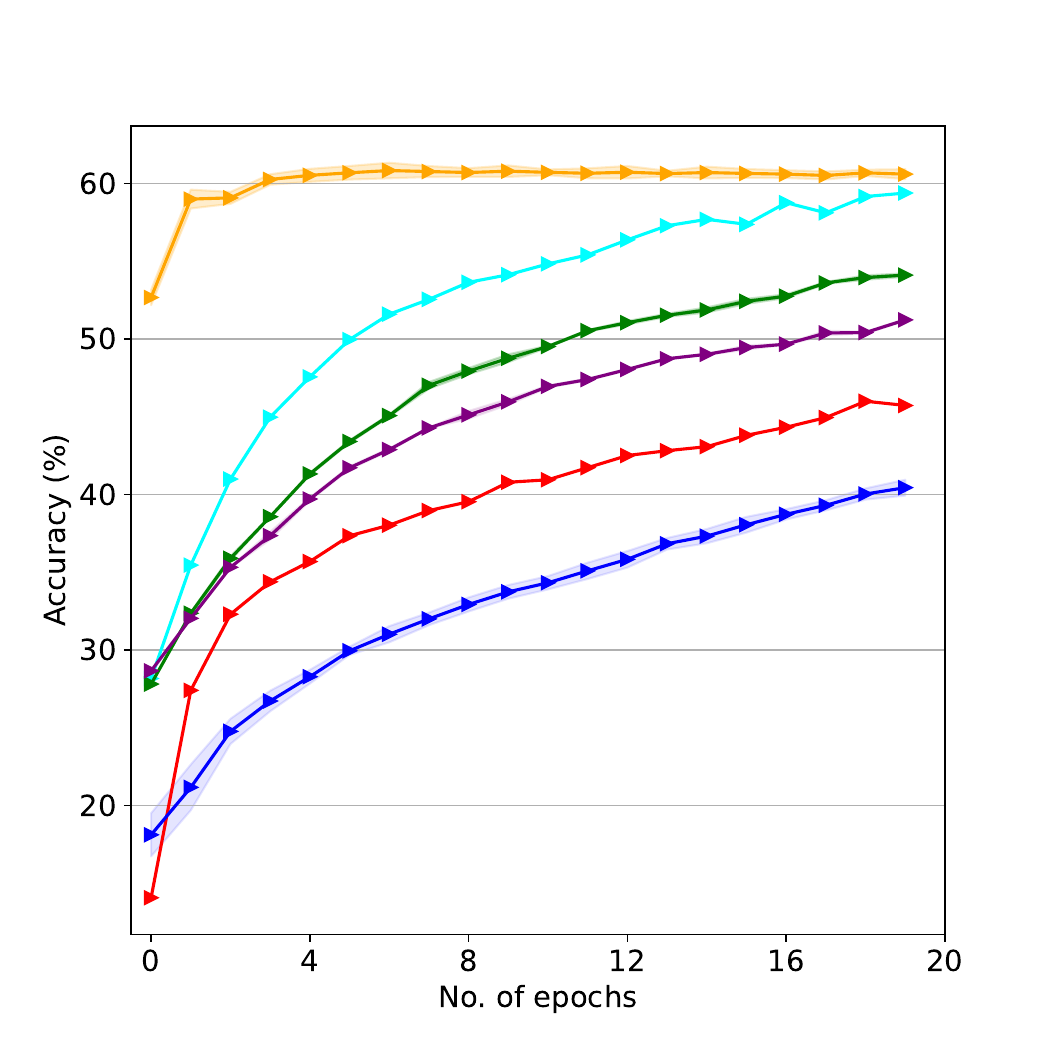}
\centerline{(a) Using 20\% of training set and 20 epochs}\medskip
\end{minipage}
\begin{minipage}{7cm}
\centering
\includegraphics[width=5cm]{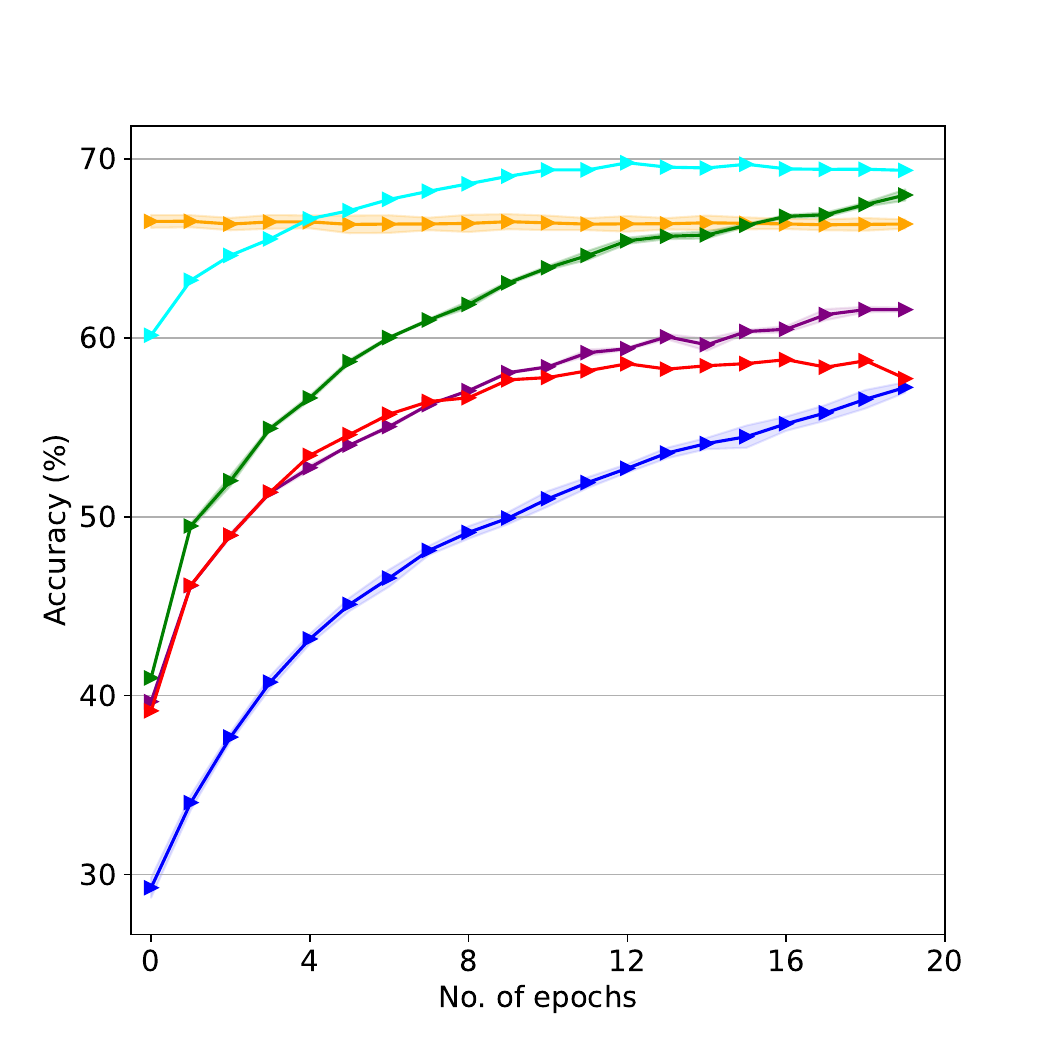}
\centerline{(b) Using 100\% of training set and 20 epochs}\medskip
\end{minipage}
\begin{minipage}{1.8cm}
\centering
\centerline{}\medskip
\end{minipage}
\begin{minipage}{7cm}
\centering
\includegraphics[width=5cm]{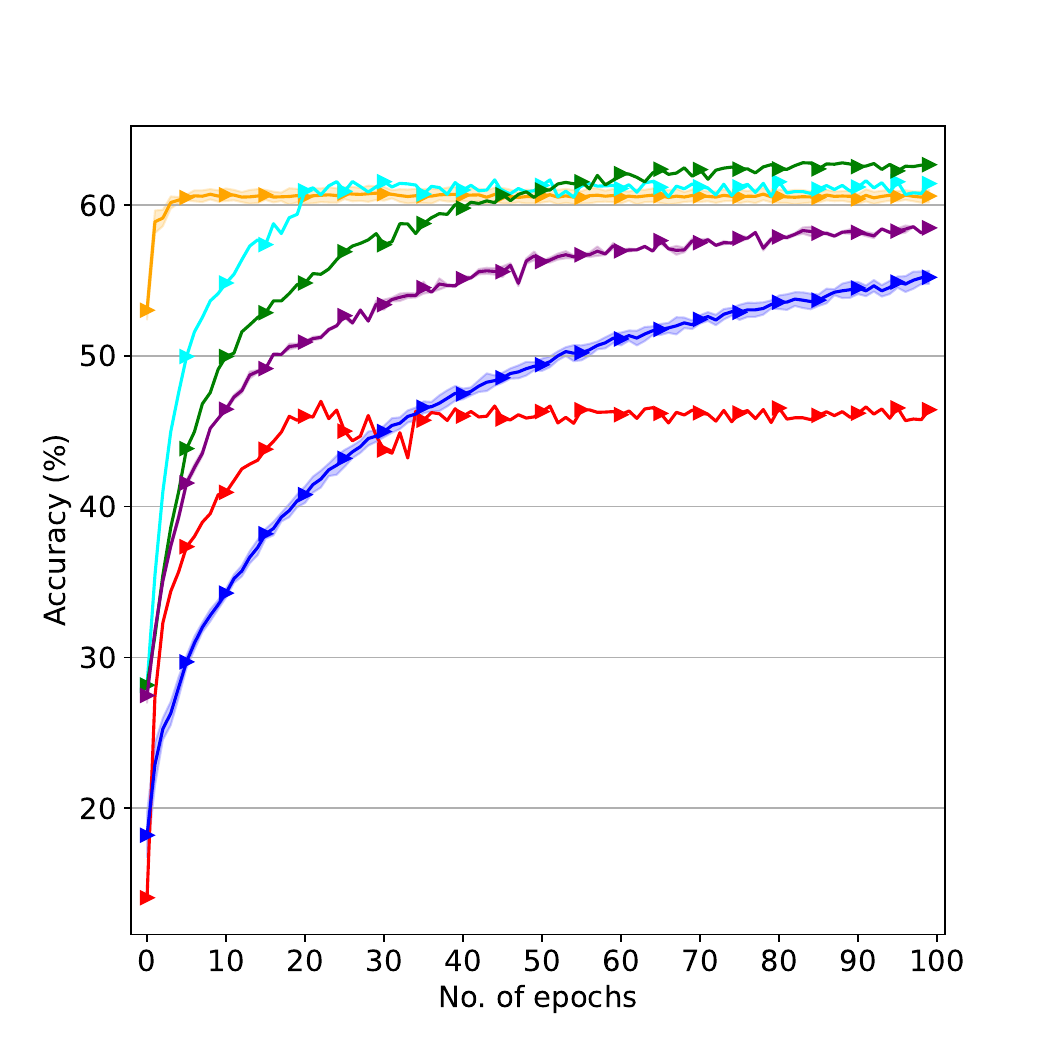}
\centerline{(c) Using 20\% of training set and 100 epochs}\medskip
\end{minipage}
\begin{minipage}{7cm}
\centering
\includegraphics[width=5cm]{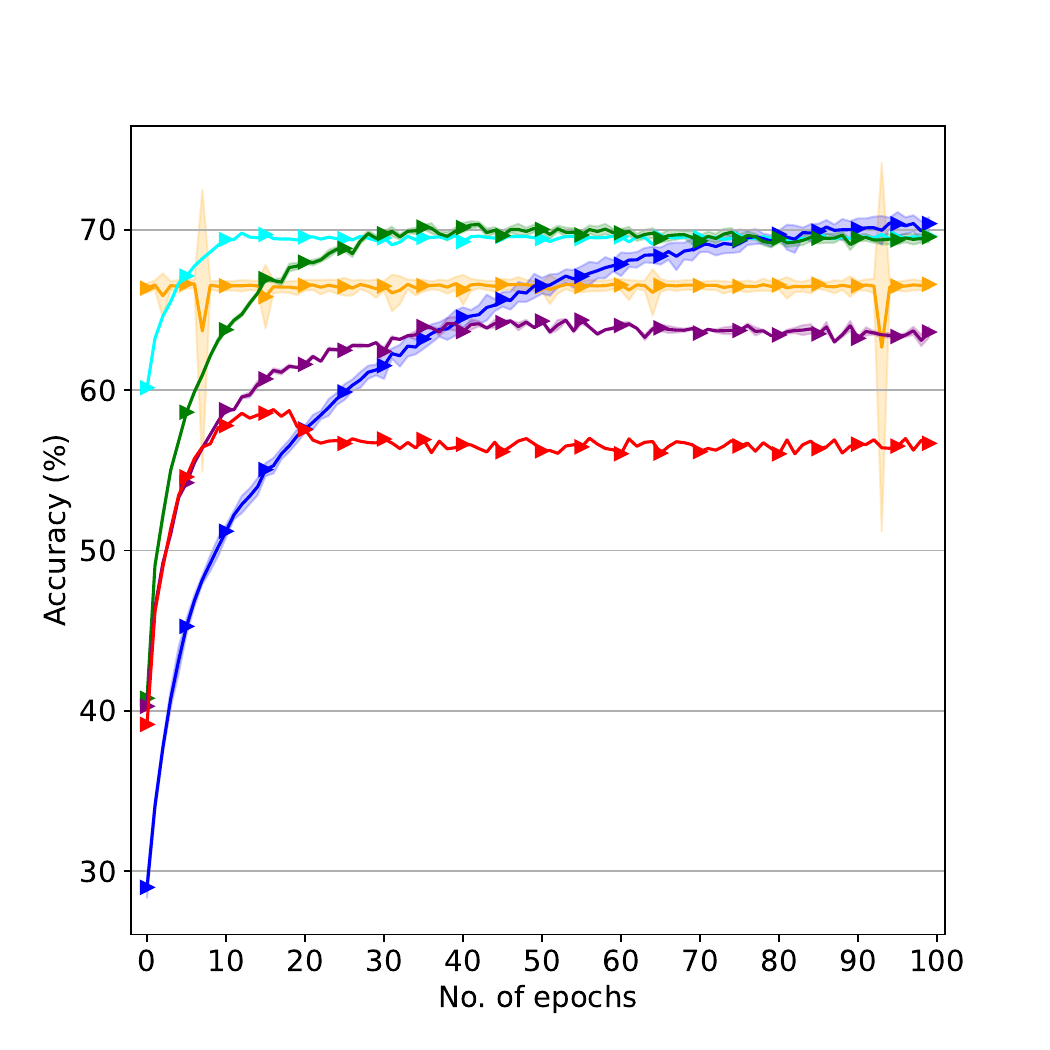}
\centerline{(d) Using 100\% of training set and 100 epochs}\medskip
\end{minipage}
\begin{minipage}{1.8cm}
\centering
\includegraphics[width=1.8cm]{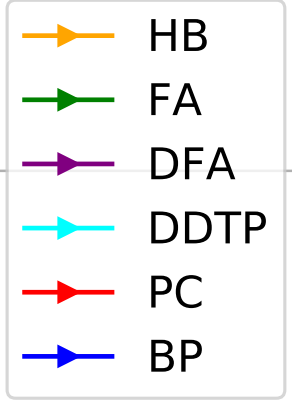}
\centerline{}\medskip
\end{minipage}
\caption{Experiments using limited data or limited training budget or both. Bio-algorithms perform much better than BP with Hebbian in particular surpassing BP by 20\% for the case of lesser data and fewer epochs (Fig. (a)). Hebbian also converges in around 5 epochs. Using CIFAR-10 dataset. Mean values plotted and standard deviation denoted by shaded regions over 10 runs.}
\label{fig:less_data_and_fewer_epochs}
\end{figure*}

Hebbian learning has been around since 1949 when it was first proposed by Donald Hebb \cite{Brain_Function}. It has since been applied in many different areas including continual learning \cite{pmlr-v80-miconi18a, thangarasa2020enabling}, principal components analysis \cite{Oja1982SimplifiedNM, Sanger, Gorrell2006GeneralizedHA} and biology \cite{Oja1989NeuralNP, Keysers, Thiagarajan} amongst others. Hebbian learning has recently been adapted for modern neural networks and applied to train deep learning models \cite{Krotov7723, Amato2019HebbianLM, 9207242, hebbnet, miconi}. However, the focus for most of these studies has been to get Hebbian learning successfully running on deep learning architectures. A gap remains in benchmarking Bio-learning vs. Backprop on multiple aspects like training with fewer epochs, utilizing less data, evaluating the performance under network sparsity, etc. We focus on these aspects and present novel findings on the performance of Bio-learning vs. Backprop in this work. 

Another Bio-inspired algorithm that has become popular is Feedback Alignment \cite{Lillicrap2016RandomSF}. It works by having a different set of weights for the forward and backward pass. In particular, the weights for the backward pass are a set of fixed random matrices over which the gradients are back-propagated. So, it is similar to BP in that an error is still calculated and back-propagated but it is different in that the weights for the backward pass do not have to be the same as the forward pass. This is biologically desirable because it breaks the symmetry condition requiring the forward synapses to also act as backward synapses. 

A variant of FA, Direct Feedback Alignment (DFA) \cite{DFA}, relaxes this condition further whereby the error does not need to be back-propagated across every layer but can directly flow from the last layer to the layer in question through a shortcut connection. This makes the model even more biologically plausible and has also seen great success with many works derived from it \cite{10.3389/fnins.2019.00525, 10.5555/3495724.3496508, scaling_dfa}. 

The fourth broad category of Bio-algorithms corresponds to Target Propagation algorithms that include Target Propagation \cite{TargetProp}, Difference Target Propagation \cite{theoretical_framework}, Equilibrium Propagation \cite{EquiProp}, amongst others. Target Propagation (TP) tries to solve the problem of credit assignment and computes targets rather than gradients at each layer. A variant of TP, Difference Target Propagation (DTP) actually works well in practice and differs from TP in that it linearly corrects for the imperfectness of TP. Equilibrium Propagation (EP) is similar in idea but is used for energy based networks like Hopfield networks. 

Lastly, other Bio-algorithms include Predictive Coding \cite{PC}, Direct Random Target Projection \cite{10.3389/fnins.2021.629892}, Weight Mirroring \cite{WM}, Local Representation Alignment \cite{LRA}, amongst others. Predictive coding is a theoretical neuroscience inspired algorithm approximating BP using local learning rules. It attempts to arrive at similar gradient updates as BP but using only biologically plausible local updates. It has gained traction in recent years and presents a promising new direction for bridging Bio-learning and BP.  

We find that while many interesting Bio-algorithms have been proposed in the literature, a thorough benchmarking between BP and various Bio-algorithms has not been undertaken on aspects outside the standard measurement metrics like test accuracy \cite{bartunov2018, xiao2018}. Some papers propose theoretical frameworks for comparison \cite{cricket} or provide brief comments on other aspects, for instance, number of training episodes \cite{BackpropFreeRL}, but do not conduct in-depth experiments and ablations tantamount to a usable research benchmark. In particular, areas of i) Limited training data ii) Limited training epochs and iii) Fewer network parameters have not been benchmarked before in an in-depth manner. We bridge this gap and benchmark five popular and varied Bio-algorithms, namely, Hebbian Learning, FA, DFA, DTP and Predictive coding to give a wide-ranging coverage of the various groups of Bio-algorithms. We conduct experiments on multiple datasets in carefully setup experiments, to demonstrate empirically the areas where Bio-algorithms surpass BP and can be used by practitioners outside the Bio-learning community for their specific deep learning tasks.

\begin{figure*}[h]
\centering
\begin{minipage}{7cm}
\centering
\includegraphics[width=5cm]{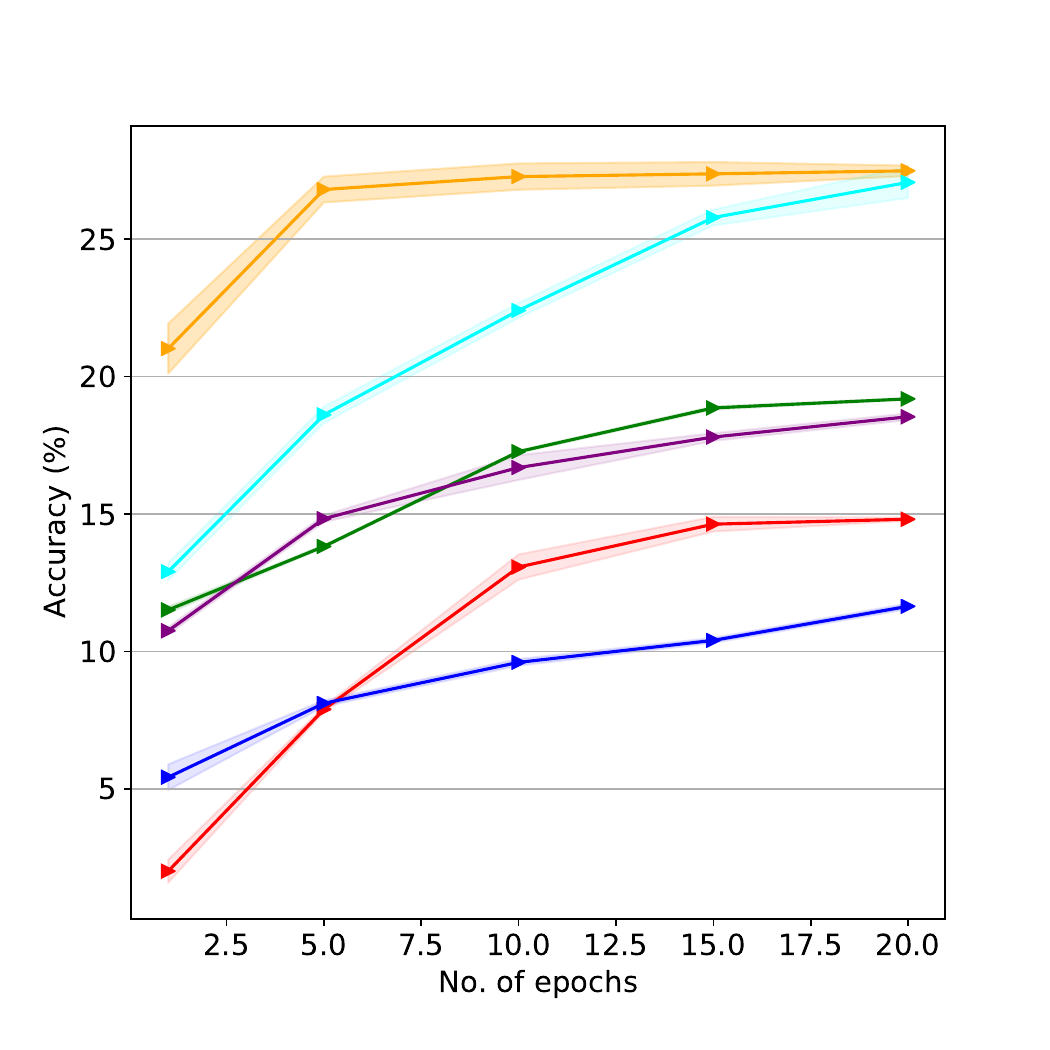}
\centerline{(a) Using 20\% of training set and 20 epochs}\medskip
\end{minipage}%
\begin{minipage}{7cm}
\centering
\includegraphics[width=5cm]{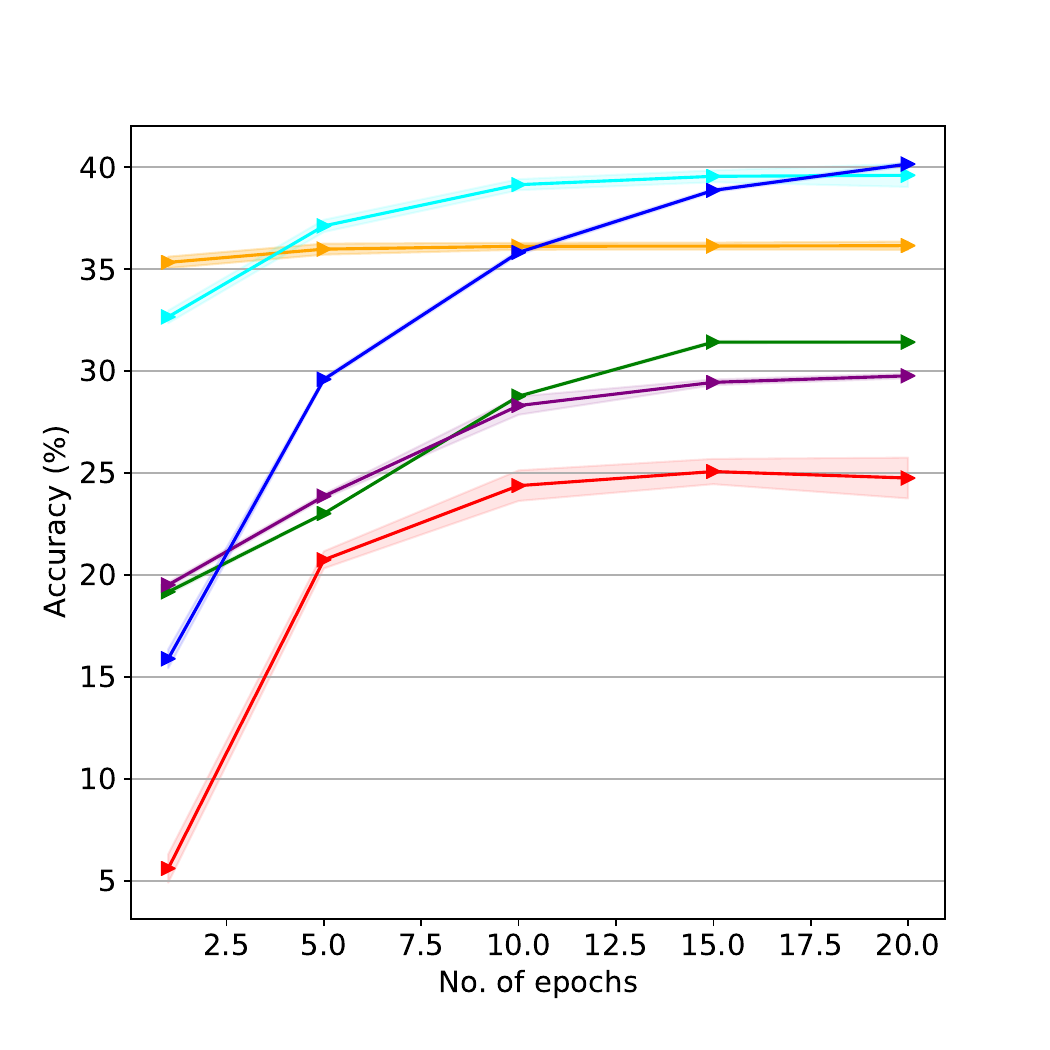}
\centerline{(b) Using 100\% of training set and 20 epochs}\medskip
\end{minipage}
\begin{minipage}{1.8cm}
\centering
\centerline{}\medskip
\end{minipage}
\begin{minipage}{7cm}
\centering
\includegraphics[width=5cm]{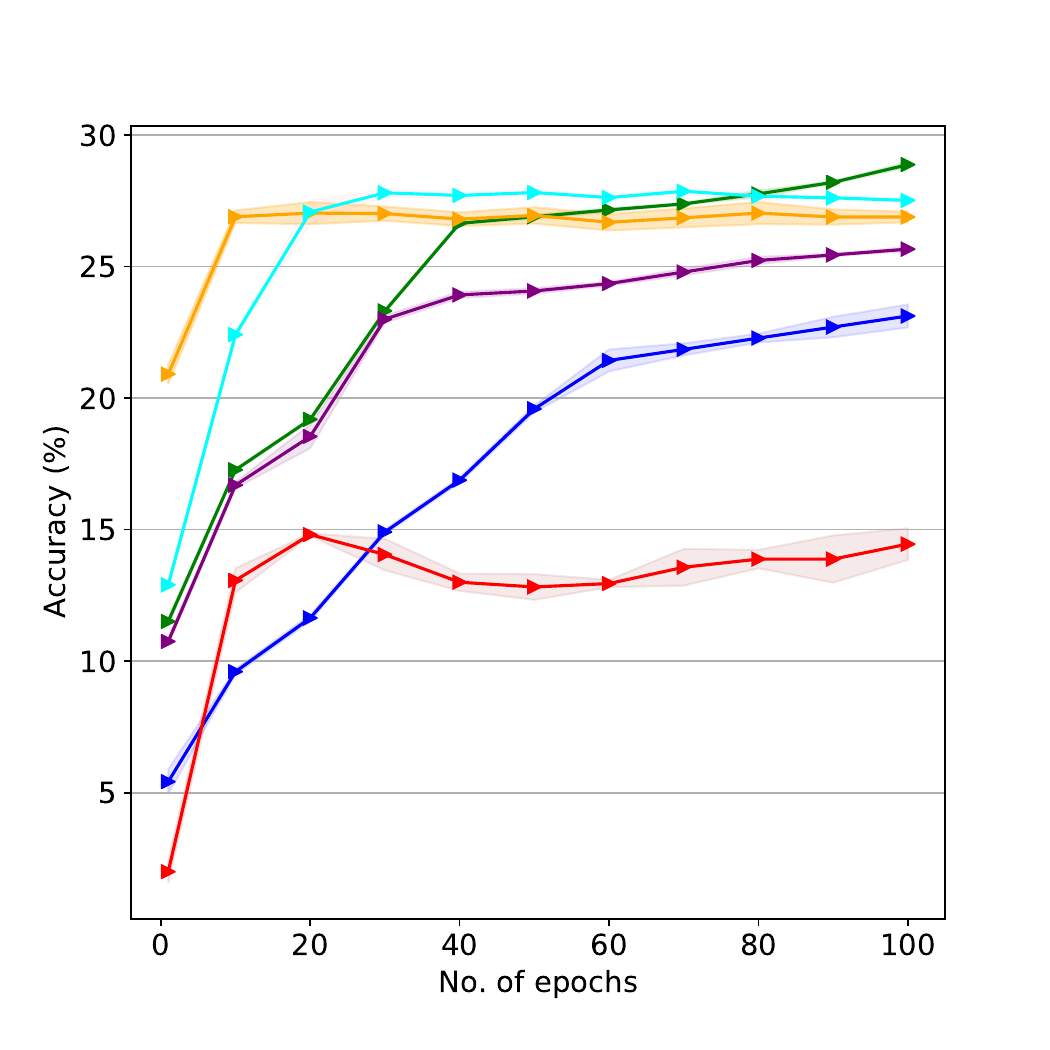}\textbf{s}
\centerline{(c) Using 20\% of training set and 100 epochs}\medskip
\end{minipage}%
\begin{minipage}{7cm}
\centering
\includegraphics[width=5cm]{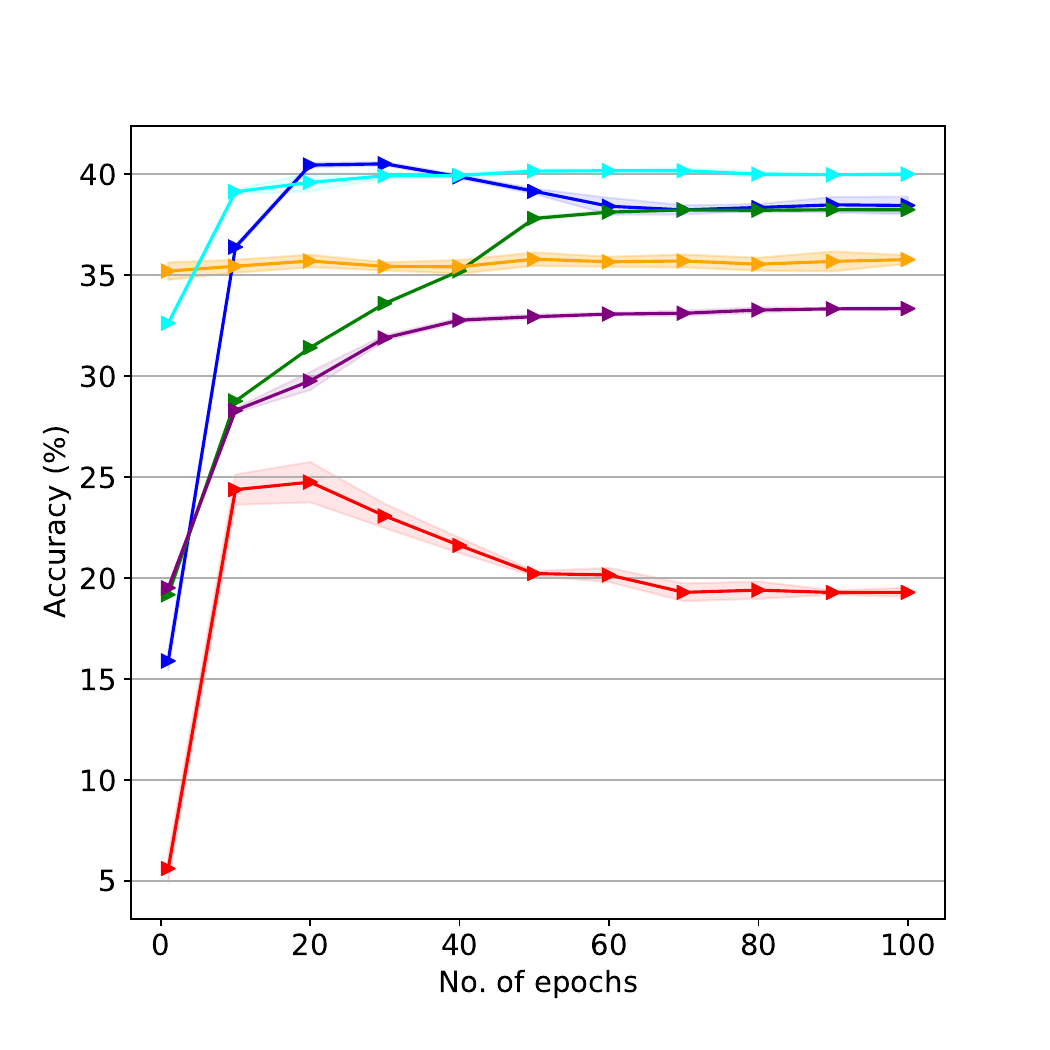}
\centerline{(d) Using 100\% of training set and 100 epochs}\medskip
\end{minipage}
\begin{minipage}{1.8cm}
\centering
\includegraphics[width=1.8cm]{Figures/cifar100-legend-CNN.png}
\centerline{}\medskip
\end{minipage}
\caption{Experiments using limited data or limited training budget or both. Bio-algorithms perform much better than BP with Hebbian in particular surpassing BP by 16\% for the case of lesser data and fewer epochs (Fig. (a)). Hebbian also converges in around 5 epochs. Using CIFAR-100 dataset. Mean values plotted and standard deviation denoted by shaded regions over 10 runs.}
\label{fig:less_data_and_fewer_epochs_cifar100}
\end{figure*}

\begin{figure}[h]
\begin{minipage}{6cm}
\centering
\includegraphics[width=6cm]{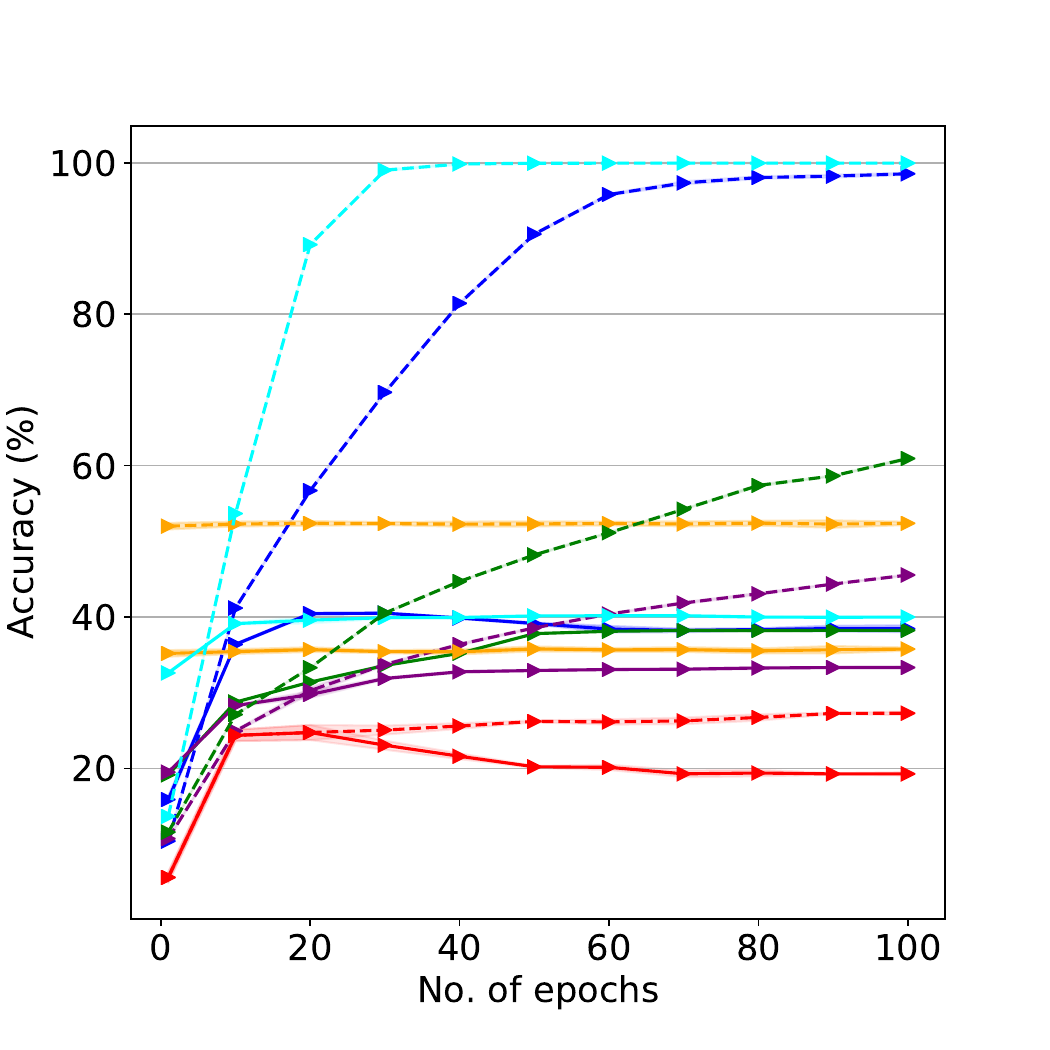}
\end{minipage}
\begin{minipage}{1.7cm}
\includegraphics[width=1.7cm]{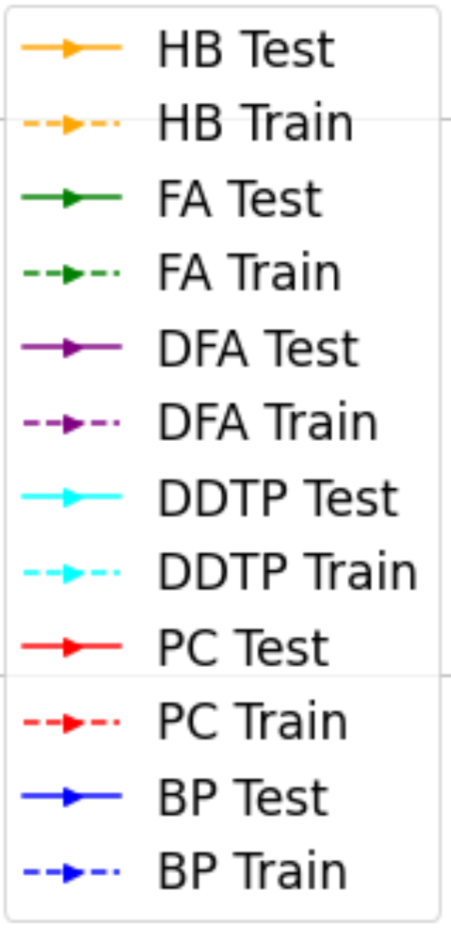}
\end{minipage}
\caption{Training and test accuracies for various algorithms on CIFAR-100 dataset using 100\% of the training set and 100 epochs. BP and DDTP substantially over-fit on the training data (around 60\% for BP). The other Bio-algorithms (HB, FA, DFA, PC) perform much better on over-fitting with training accuracy being close to test accuracy (ranging around 6\%-22\%).}
\label{fig:training}
\end{figure}

\begin{figure*}[h]
\centering
\begin{minipage}{6cm}
\centering
\includegraphics[width=5cm]{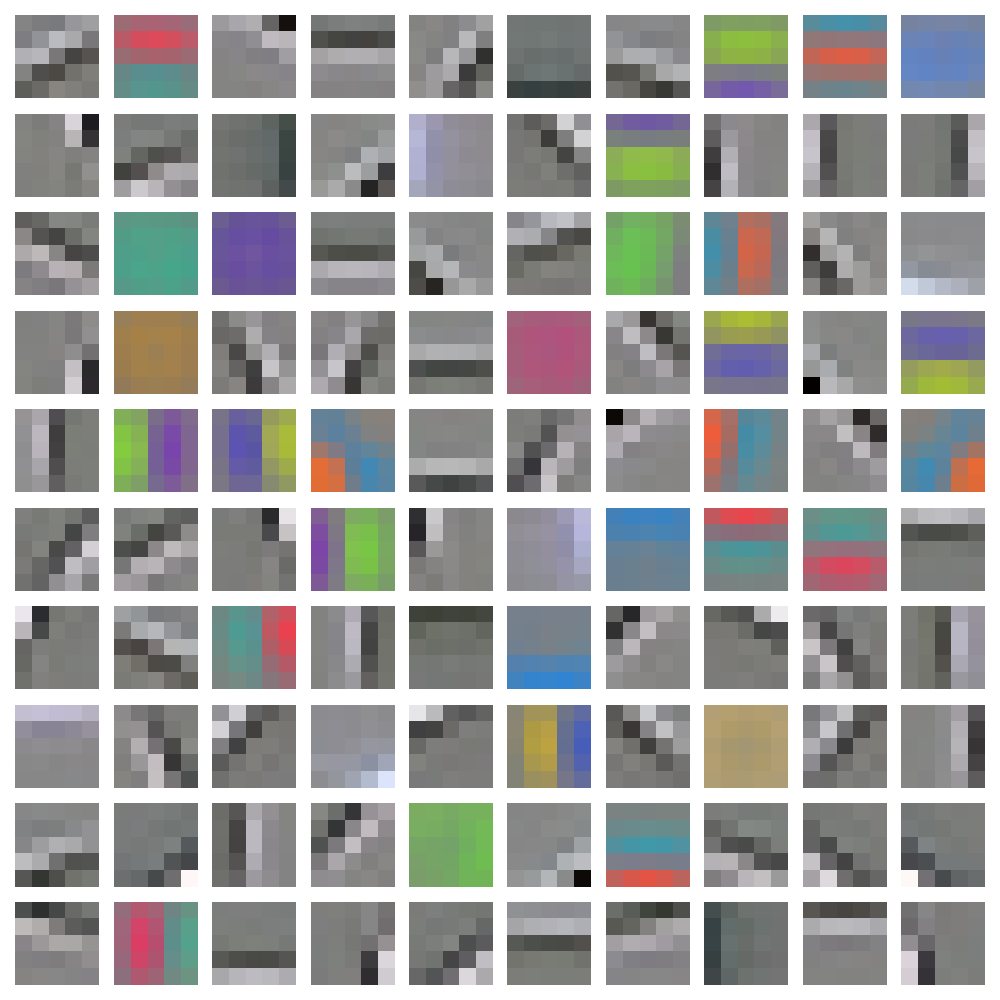}
\centerline{(a) Filters learnt by Hebbian learning}\medskip
\end{minipage}%
\hspace{1cm}
\begin{minipage}{6cm}
\centering
\includegraphics[width=5cm]{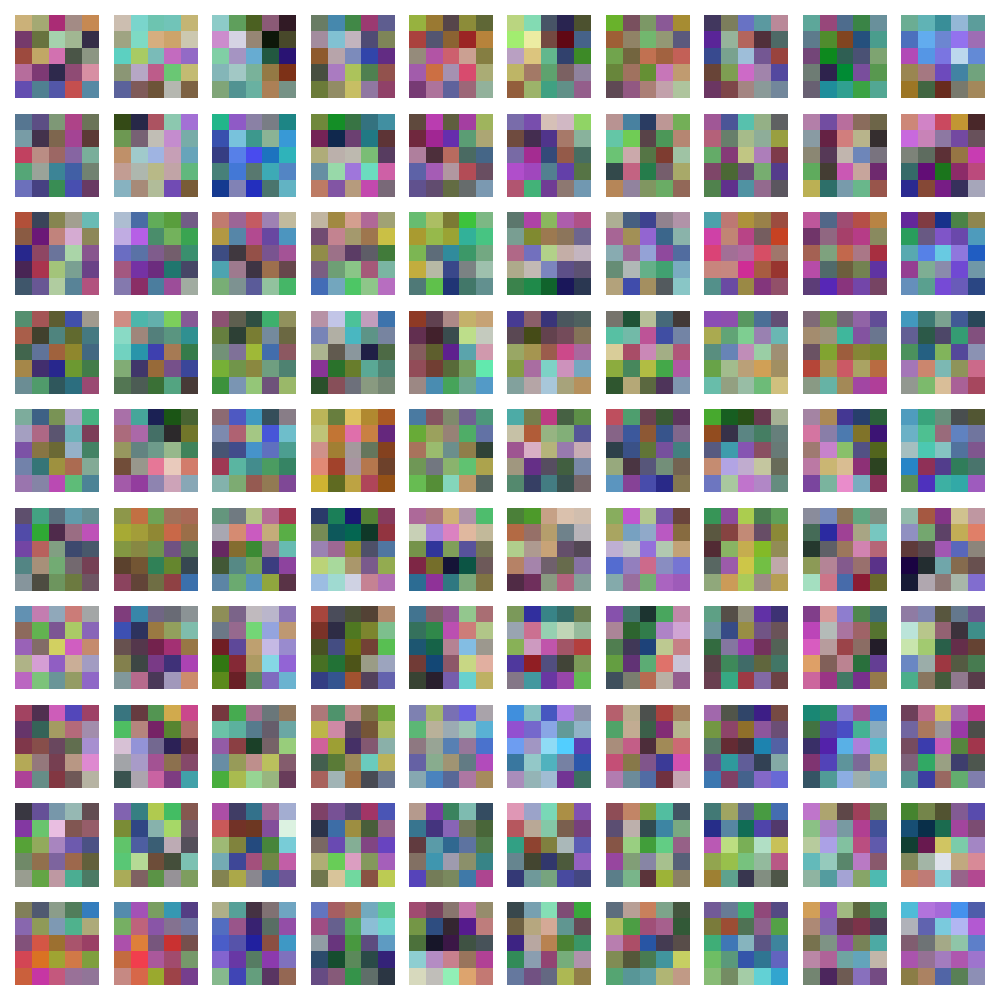}
\centerline{(b) Filters learnt by Backprop}\medskip
\end{minipage}
\caption{Visualizing the filters learnt by Hebbian learning vs. Backprop. Hebbian learning learns much more fine-grained representations compared to Backprop. Hebbian learns filters which are orientation-sensitive (vertical, horizontal and diagonal grayscale edges), color-sensitive (blue, green, red filters) or both (filters with combination of orientation and two or more colors). Plotting filters from the first CNN layer on the CIFAR-10 dataset after 100 epochs of training.}
\label{fig:representations}
\end{figure*}

\begin{figure*}[!h]
\begin{minipage}{.24\textwidth}
    \centering
    \includegraphics[width=\textwidth]{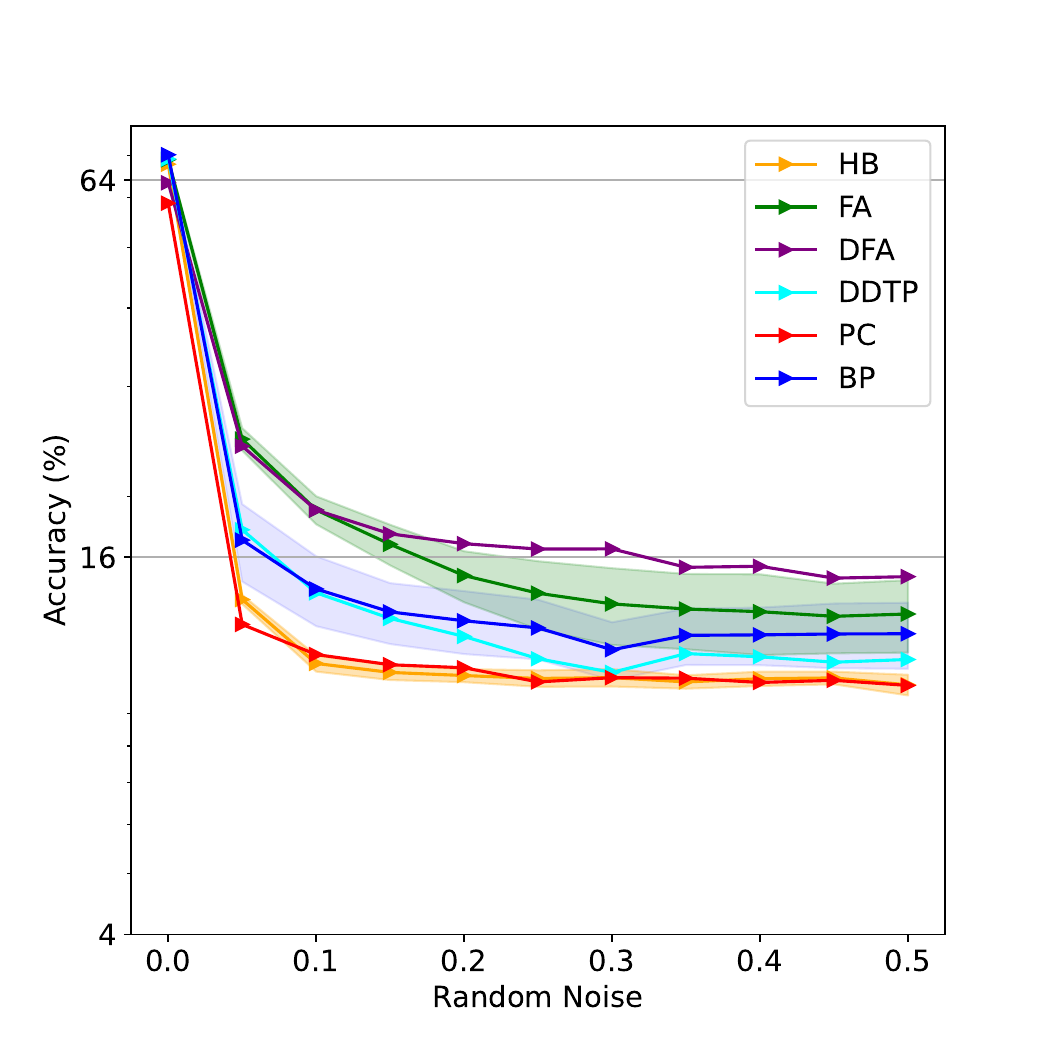}
    \centerline{(a)}\medskip
\end{minipage}
\begin{minipage}{.24\textwidth}
    \centering
    \includegraphics[width=\textwidth]{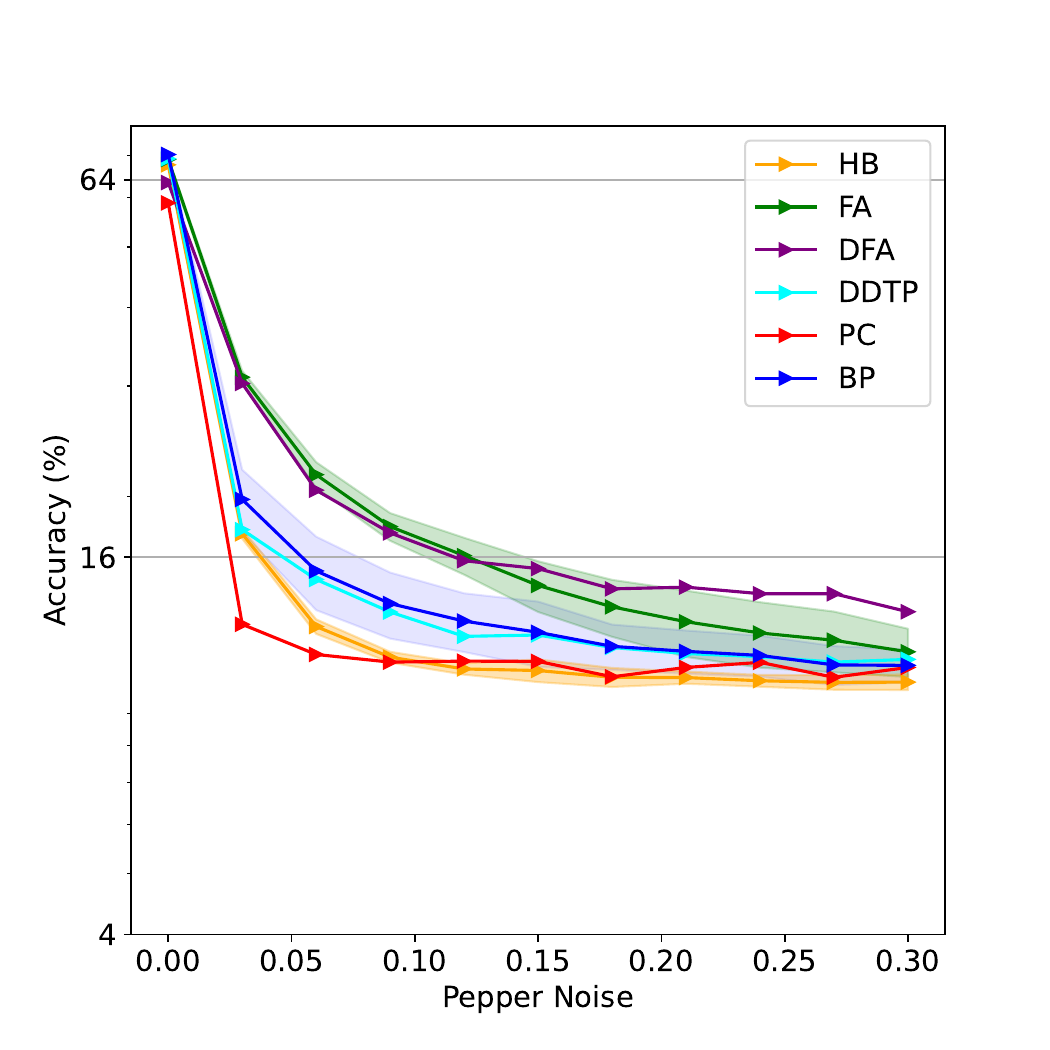}
    \centerline{(b)}\medskip
\end{minipage}
\begin{minipage}{.24\textwidth}
    \centering
    \includegraphics[width=\textwidth]{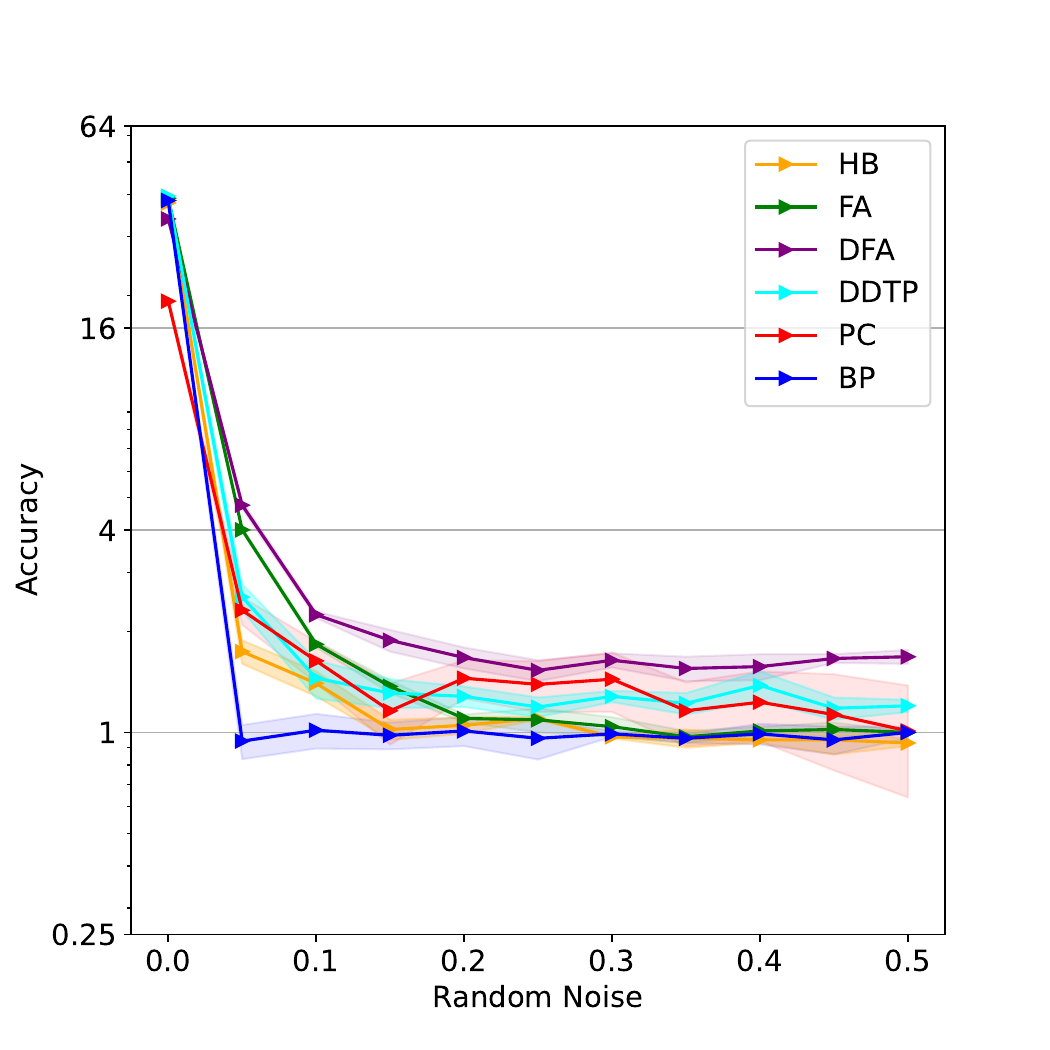}
    \centerline{(c)}\medskip
\end{minipage}
\begin{minipage}{.24\textwidth}
    \centering
    \includegraphics[width=\textwidth]{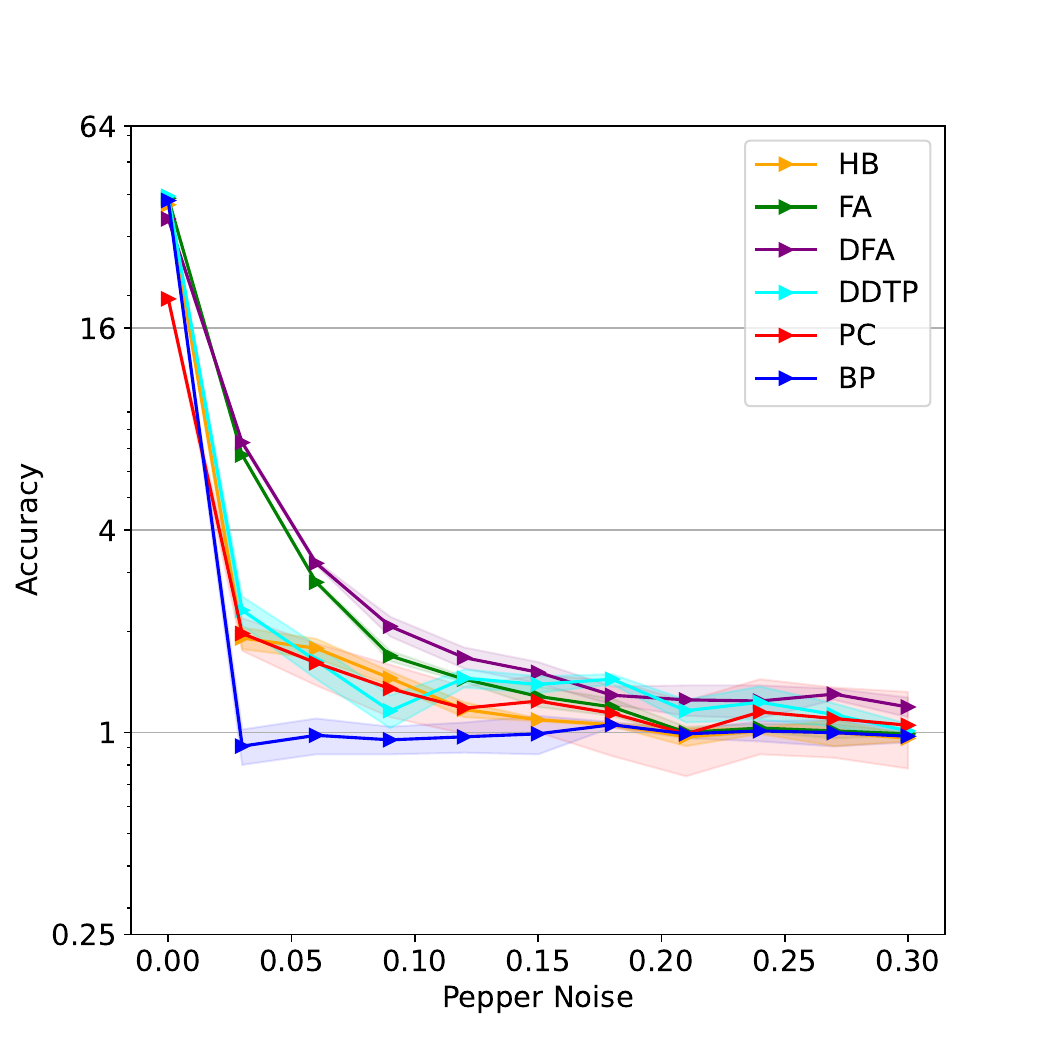}
    \centerline{(d)}\medskip
\end{minipage}
    \caption{Experiments on adding Random noise and Pepper noise to the input images. Using CIFAR-10 dataset for Fig. (a) \& (b) and CIFAR-100 for Fig. (c) \& (d). Bio-learning (especially DFA) is quite robust to both Random and Pepper noise on both CIFAR-10 and CIFAR-100 datasets and surpasses BP. Mean values plotted and standard deviation denoted by shaded region over 10 runs.}
    \label{fig:noise}
\end{figure*}

\begin{figure*}[h]
\centering
\begin{minipage}{7cm}
\centering
\includegraphics[width=5cm]{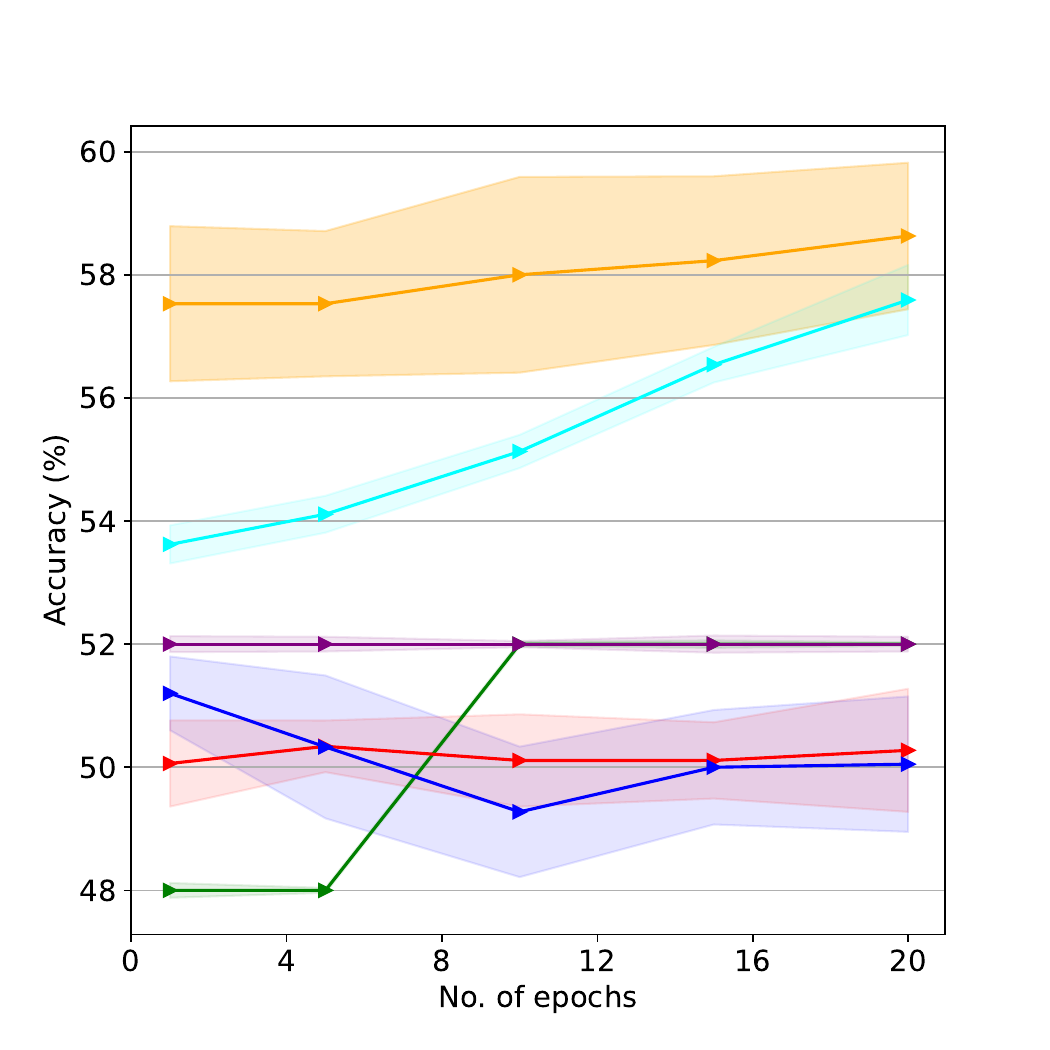}
\centerline{(a) Using 20\% of training set and 20 epochs}\medskip
\end{minipage}%
\begin{minipage}{7cm}
\centering
\includegraphics[width=5cm]{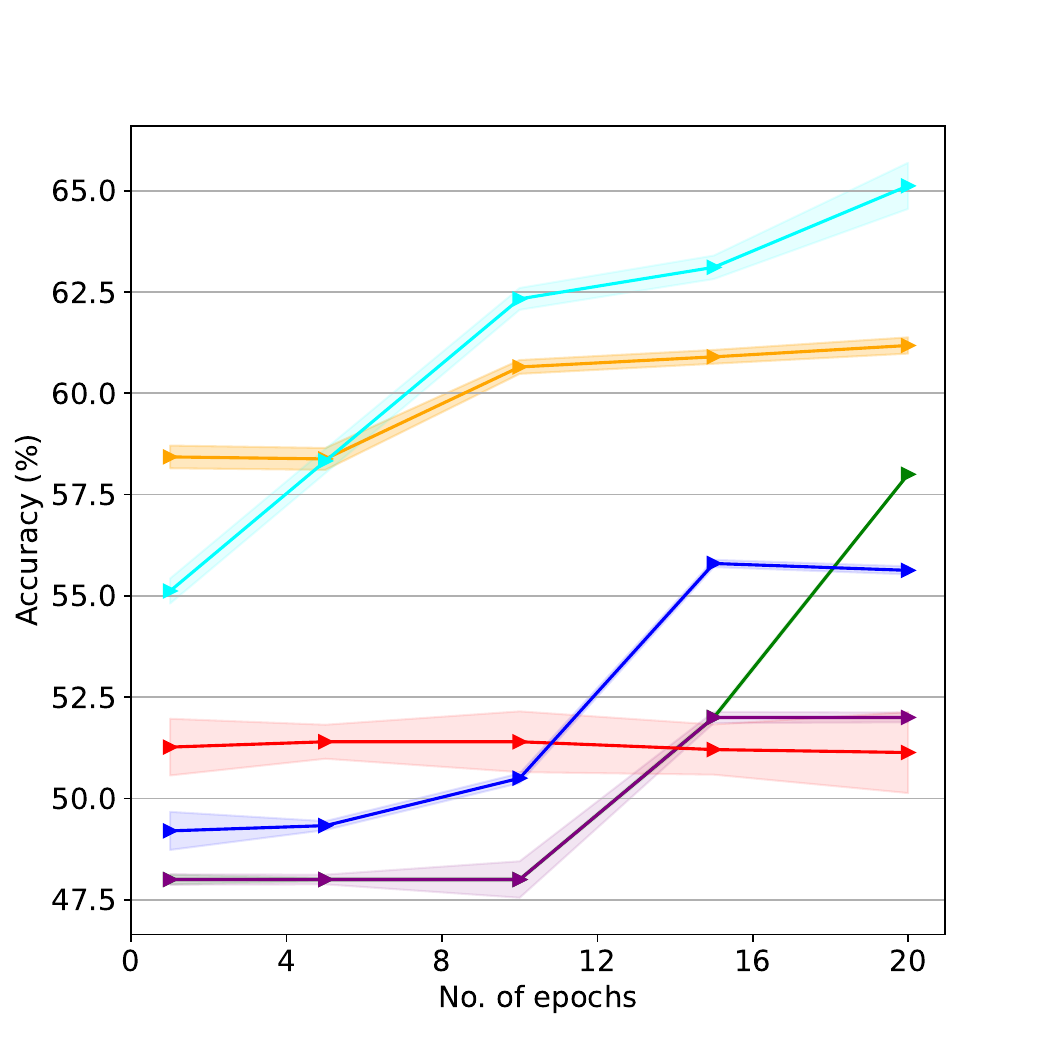}
\centerline{(b) Using 100\% of training set and 20 epochs}\medskip
\end{minipage}
\begin{minipage}{1.8cm}
\centering
\centerline{}\medskip
\end{minipage}
\begin{minipage}{7cm}
\centering
\includegraphics[width=5cm]{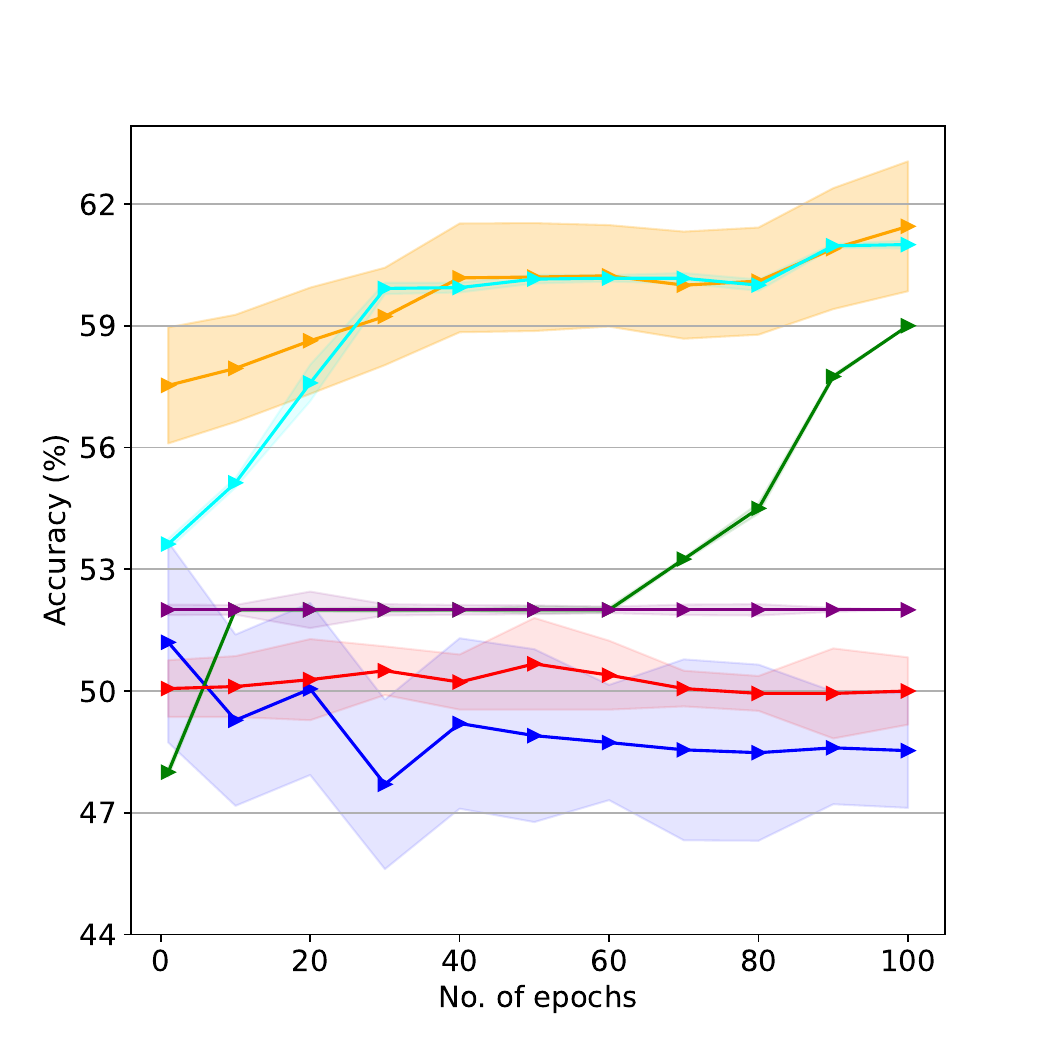}
\centerline{(c) Using 20\% of training set and 100 epochs}\medskip
\end{minipage}%
\begin{minipage}{7cm}
\centering
\includegraphics[width=5cm]{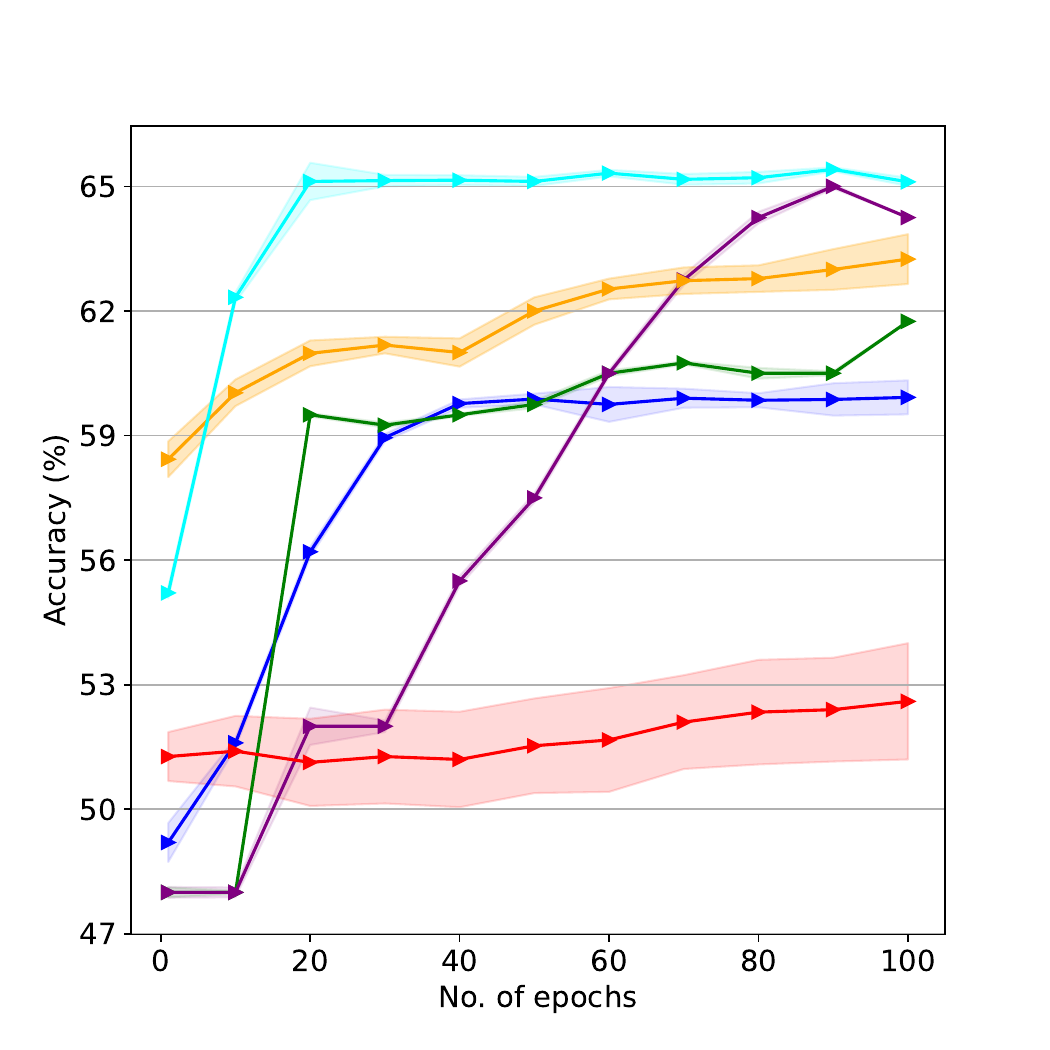}
\centerline{(d) Using 100\% of training set and 100 epochs}\medskip
\end{minipage}
\begin{minipage}{1.8cm}
\centering
\includegraphics[width=1.8cm]{Figures/cifar100-legend-CNN.png}
\centerline{}\medskip
\end{minipage}
\caption{Experiments using limited data or limited training budget or both. Bio-algorithms perform much better than BP for all cases, with Hebbian and DDTP in particular surpassing BP by 12\% for the case of lesser data (Fig. (c)). Using Stanford Sentiment Treebank (SST-2) dataset. Mean values plotted and standard deviation denoted by shaded regions over 10 runs.}
\label{fig:less_data_and_fewer_epochs_sst2}
\end{figure*}
\section{Approach}

In this section, we introduce the three Bio-inspired learning algorithms used in our experiments, that is, Hebbian learning (Grossberg’s Instar Rule), Feedback Alignment (FA) and Direct Feedback Alignment (DFA).

\subsection{Hebbian Learning}

\quad \textbf{Vanilla Hebbian learning:} We first describe the vanilla Hebbian rule which states that when a neuron is fired by the stimulation of another neuron connected to it, the strength of the weighted connection between the two is enhanced, and vice versa \cite{10.1007/978-3-642-70911-1_15, lowel1992selection}. Hebbian learning is a completely local, unsupervised learning algorithm and  satisfies biological plausibility. The weight update is given by

\begin{equation}
\Delta{w_{ij}} = \eta z_jx_i
\label{eq1}
\end{equation}

where $w_{ij}$ is the weight connecting neurons \textit{i} and \textit{j}, $x_{i}$ and $z_{j}$ are the input and output (after activation) of neuron \textit{j} and $\eta$ is the learning rate. 

However, this vanilla rule has a significant drawback that can cause the weight to grow infinitely. Therefore, variants of the vanilla Hebbian rule have been proposed to address this problem. 

\textbf{Grossberg’s Instar Rule:} Instar rule \cite{grossberg1976adaptive} introduces the mechanism of weight decay for solving the problem of unbounded growth of weights. From a mathematical point of view, the introduction of the weight decay mechanism leads the weight vector $w_{ij}$ to cluster towards the input vector $x_i$ \cite{lagani2021hebbian}. Giving the same input to the neuron repeatedly will eventually make the angle between the weight vector and the input vector to be zero. In other words, the maximum inner product between $x_i$ and $w_{ij}$ will occur in the above case.

\begin{equation}
\Delta{w_{ij}} = \eta z_j(x_i-w_{ij}) \label{eq2}
\end{equation}

where the symbol interpretation is same as Equation \ref{eq1}.

\subsection{Feedback Alignment}

\quad FA was proposed to eliminate the weight transport mechanism relied on by Backprop (BP), as it is not considered biologically plausible in the brain. Weight transport implies that the weight matrix between neurons needs to be involved in operations for both forward and backward paths \cite{akrout2019deep}. Unlike the BP algorithm, FA replaces the transpose of the weight matrix that is required in the backward pass, with a fixed random matrix instead \cite{Lillicrap2016RandomSF}.

In the forward pass, the feed-forward calculation follows:

\begin{equation}
a_{i+1} = w_{i+1}z_i+b_{i+1} \label{eq3}
\end{equation}

\begin{equation}
z_{i+1} = \sigma (a_{i+1}) \label{eq4}
\end{equation}

where $z_i$ is the output of the $i$-th layer, $w_{i+1}$ is the feed-forward weight, $b_{i+1}$ is the bias vector, and $\sigma$ is the non-linear activation function.

In the backward pass of FA, the error $e$ follows:

\begin{equation}
e_{i} = (B_{i+1}e_{i+1}) \odot \sigma'(a_i) \label{eq5}
\end{equation}

where $B_{i+1}$ is the fixed random matrix to pass the error through, $\odot$ is an element-wise multiplication operator and $\sigma'$ is the derivative of the non-linear activation function.

Thus, the weight update becomes

\begin{equation}
\Delta{w_{i+1}} = -\eta e_{i+1} z_{i}^{T}
\label{eq6}
\end{equation}

\subsection{Direct Feedback Alignment}

\quad DFA is a modified version of FA that breaks the mechanism whereby errors need to be propagated layer by layer. Instead, it passes the errors directly from the final layer to the layer in question, circumventing any intermediate layers \cite{DFA}. This allows the algorithm to have greater biological plausibility.

The backward pass of DFA is slightly different from FA and follows:

\begin{equation}
e_{i} = (B_{i+1}e_{f}) \odot \sigma'(a_i) \label{eq7}
\end{equation}

where $e_{f}$ is the error after the final layer.

The weight update for DFA remains the same as that for for FA, and follows Equation \ref{eq6}.
\section{Experiments}
\label{results}

We conduct experiments on Hebbian Learning (HB), Feedback Alignment (FA), Direct Feedback Alignment (DFA), Difference Target Propagation (DTP), Predictive Coding (PC) and Backprop (BP). Multiple datasets including CIFAR-10 \cite{Krizhevsky09learningmultiple}, CIFAR-100 \cite{Krizhevsky09learningmultiple} and Stanford Sentiment Treebank (SST-2) \cite{sst2} are used for testing. We use Convolutional Neural Networks (CNNs) for the experiments owing to their higher accuracies than Fully Connected Networks (FCNs). For all experiments, 4 layered networks are used with the first 3 layers being convolutional layers having 100, 196 and 400 neurons respectively and the last layer being a linear layer for the case of BP, FA, DFA, DDTP and PC and ridge classifier for the case of Hebbian. For FA and DFA the implementation from \cite{10.3389/fnins.2021.629892} is used and for Hebbian learning the implementation from \cite{miconi} is used. For DTP the implementation from \cite{theoretical_framework} is used, called DDTP, while for PC the implementation from \cite{PC} is used. For fair comparison between Bio-learning and BP, vanilla networks are used for all experiments without drop-out, batch-norm or any other training tricks. Hyper-parameters used are provided in the Appendix. All experiments are conducted for 100 epochs unless stated otherwise.

We first conduct experiments on the limited data and limited training budget scenarios where only a subset of the original dataset or a smaller training budget is available for training (Fig. \ref{fig:less_data_and_fewer_epochs} (a)). As can be seen, all the five Bio-learning algorithms perform much better than BP. Hebbian learning, in particular, performs really well achieving 20\% higher accuracy than BP for the case of less data and fewer training epochs. Also, notably, Hebbian learning converges much faster than BP, usually achieving convergence in just around 5 epochs. A similar trend is seen when the epochs is increased to 100 (Fig. \ref{fig:less_data_and_fewer_epochs} (c)), where FA achieves the top accuracy amongst all the algorithms. When the condition of less data availability is relaxed (Fig. \ref{fig:less_data_and_fewer_epochs} (b)), even then Bio-algorithms surpass BP by a sizable margin. Only in the case of availability of the full dataset and the full training budget (Fig. \ref{fig:less_data_and_fewer_epochs} (d)), is BP able to get comparable accuracy as the top performing Bio-algorithm. Even then BP requires the full 100 epochs to learn whereas all the five Bio-algorithms finish learning within 40 epochs and within 5 epochs for Hebbian learning.

Similar results are seen for the case of CIFAR-100 (Fig. \ref{fig:less_data_and_fewer_epochs_cifar100}). Hebbian learning (HB) and Difference Target Propagation (DTP) perform the best and much better than BP (around 16\% better in terms of accuracy) for the case of limited data and limited training budget (Fig. \ref{fig:less_data_and_fewer_epochs_cifar100} (a)). Even with increasing the training budget, Bio-algorithms still surpass BP (Fig. \ref{fig:less_data_and_fewer_epochs_cifar100} (c)). Only when the full dataset is shown to BP is it able to achieve comparable accuracy to the top performing Bio-algorithm DTP (Fig. \ref{fig:less_data_and_fewer_epochs_cifar100} (b)). DTP also outperforms BP for the case of the full dataset and full training budget (Fig. \ref{fig:less_data_and_fewer_epochs_cifar100} (d)). Notably, Hebbian learning continues to learn extremely fast and converges to stable accuracy within 5 epochs where other algorithms take around 60 epochs to converge. We also plot the training curves for the algorithms to analyse their over-fitting behavior, Figure \ref{fig:training}. We find that BP and DDTP substantially over-fit on the training data (around 60\% for the case of BP). The other Bio-algorithms (HB, FA, DFA, PC) however, perform much better on over-fitting with training accuracy being close to test accuracy (ranging around 6\%-22\%).

To understand the superior performance of Hebbian learning, we visualize the features learnt by Hebbian learning vs. BP. Referring to Fig. \ref{fig:representations}, we note that the filters learnt by Hebbian learning are much more fine-grained compared to BP. Hebbian learns filters which are orientation-sensitive (vertical, horizontal and diagonal grayscale edges), color-sensitive (blue, green, red filters) or both (filters with combination of orientation and two or more colors). BP on the other hand seems to learn more random-like features with no clear orientation or color-sensitivity. We believe this fine-grained representation ability of Hebbian might be one of the underlying reasons for its better performance and more research can be conducted on this to provide conclusive evidence.

We also note that Bio-learning is quite robust to different kinds of noises. Referring to Fig. \ref{fig:noise}(a) and Fig. \ref{fig:noise}(b), we see that Bio-learning (especially DFA) performs better than Backprop under the addition of random noise and pepper noise, respectively for CIFAR-10. For CIFAR-100, DFA performs the best and even better than BP (Fig. \ref{fig:noise}(c) and Fig. \ref{fig:noise}(d)). Bio-learning is also robust to network sparsification achieving similar results as Backprop even at very high sparsity rates of 95\% of all weights pruned, see Table \ref{table:sparsity}. Using 100\% of the training set for 100 epochs for both the noise and sparsity experiments. See Appendix for the detailed procedure on network sparsification. We don't include Hebbian learning for sparsity, since as per the implementation of Hebbian learning \cite{miconi}, it is already pruned by default and hence, we are unable to perform a fair comparison with the other networks. 

Lastly, to validate the key findings of Bio-algorithms of learning with limited data and learning with limited training budget on a completely different domain, we test the algorithms on a NLP dataset - the Stanford Sentiment Treebank (SST-2) \cite{sst2} binary sentiment classification task that classifies text-based movie-reviews into two categories, positive and negative. As can be seen from Fig. \ref{fig:less_data_and_fewer_epochs_sst2}, the findings hold for this dataset as well. Referring to Fig. \ref{fig:less_data_and_fewer_epochs_sst2}(a), Bio-algorithms surpass BP and Hebbian performs the best for the case of limited data and limited training budget (around 8\% better than BP). On increasing the training budget, Bio-algorithms still surpass BP by a significant margin (12\% for Hebbian and DDTP)(Fig. \ref{fig:less_data_and_fewer_epochs_sst2} (c)). On showing the full dataset, BP learns better and is able to improve its accuracy, although it is still surpassed by Hebbian, FA and DDTP (Fig. \ref{fig:less_data_and_fewer_epochs_sst2} (b)). For the case of the full dataset and full training budget (Fig. \ref{fig:less_data_and_fewer_epochs_sst2} (d)), BP achieves its highest accuracy but is surpassed by most of the Bio-algorithms, with DDTP performing the best. 

\begin{table}
\centering
{\begin{tabular}{|l|c|c|}
\hline
Model & CIFAR-10 (Acc.) & CIFAR-100 (Acc.) \\
\hline
FA & 62.17\% $\pm$ 0.24\% & 30.93\% $\pm$ 0.16\% \\
\hline
DFA & 57.75\% $\pm$ 0.16\% & 30.29\% $\pm$ 0.25\% \\
\hline
BP & 65.37\% $\pm$ 0.65\% & 37.68\% $\pm$ 0.48\% \\
\hline
PC & 56.2\% $\pm$ 0.48\% & 26.07\% $\pm$ 0.46\% \\
\hline
DDTP & 64.2\% $\pm$ 0.31\% & 39.07\% $\pm$ 0.22\% \\
\hline
\end{tabular}}
\caption{Experiment on the application of 95\% weight sparsity to the networks. We find that Bio-learning is robust to network sparsification. Mean and standard deviations reported over 10 runs.}
\label{table:sparsity}
\end{table}
\section{Discussion, Limitations and Future Work}
\label{Disc}

We have noted the superior performance of Bio-inspired learning compared to Backprop on several aspects. Under limited data or limited training budget Bio-inspired learning does much better and also converges much faster. This is an exciting direction and the research landscape for Bio-learning is still nascent. One of the biggest bottlenecks for Bio-learning is that the training tricks and techniques to maximize performance of Bio-learning are still under-developed. Backprop benefits from many years of combined research on optimizers, hyper-parameter tuning techniques and heuristics, custom layers like Batchnorm \cite{pmlr-v37-ioffe15} and other tricks like Dropout \cite{JMLR:v15:srivastava14a}. Such progress is severely lacking for Bio-learning making the discovery of the full potential of Bio-learning difficult. More research in these areas will help Bio-learning substantially.
\section{Conclusion}

We have investigated Bio-inspired learning algorithms like Hebbian learning, Feedback Alignment, Direct Feedback Alignment, Difference Target Propagation and Predictive Coding in this work, and benchmarked them against Backprop on multiple aspects. We measure the performance of these algorithms under conditions of i) Limited availability of training data ii) Limited training budget iii) Application of different kinds of noises on the inputs and iv) Application of parameter sparsity on the base neural network models. Hebbian learning in particular, surpasses the accuracy performance of Backprop by upto 20\% under limited data and limited training budget scenarios. Bio-algorithms also achieve much faster convergence (upto 20$\times$ for the case of Hebbian learning) compared to Backprop. These insights help motivate the usefulness of Bio-algorithms in important areas like learning with less data or learning in resource-constrained training environments like real-time devices. They highlight the need to work on Bio-learning algorithms for tangible practical benefits and beyond purely theoretical motivations. We also plot the representations learnt by the Bio-algorithms and find that they learn more fine-grained representations compared to BP. Lastly, we find that Bio-learning is robust to noise and parameter sparsity (even at extreme sparsities of 95\%), achieving similar or higher accuracy than Backprop.

{\small
\bibliographystyle{IEEEtran}
\bibliography{bib}
}

\newpage
\clearpage
\section{Appendix}
\setcounter{figure}{7} 
\setcounter{table}{1} 

\subsection{Experiment design and hyper-parameters}
We describe below in detail the hyper-parameters and the experimental setup for the various experiments conducted. For Hebbian learning, we use the same experimental setup as the original and utilize ZCA, pruning and triangle method for computing Hebbian updates. For vision datasets (CIFAR-10 and CIFAR-100), we take the hyper-parameters recommended by the original authors (for HB, FA, DFA, PC and DDTP). For BP hyper-parameters, we do a grid search to find the optimal learning rate and batch size and use the Adam optimizer. This ensures that our BP algorithm is tuned well and gives competitive performance vs. Bio-algorithms. For NLP dataset (SST-2), we do a search over the LR ($10^{-3}$ to $10^{-6}$) and Batchsize (20, 50, 100, 200) for all algorithms and choose the best performing ones. The tables below list the hyper-parameters used for each experiment.

\subsection{Network sparsification}
The procedure for network sparsification is that all methods are trained for 100 epochs on 100\% of the dataset to become fully trained. Weights are then removed as per the global magnitude pruning rule upto 95\% sparsity. The accuracy is remeasured then and reported.

\vspace{0.3cm}
\begin{table}[ht]
\centering
\small
\begin{tabular}{p{1cm}p{1.2cm}p{0.5cm}p{1.5cm}p{1.6cm}}
\hline
\multicolumn{5}{c}{\centering{Hyper-parameters for vision datasets}} \\ \hline
Algorithm & Epochs & Batch-size & Initial LR & LR Schedule\\
\hline
HB&20 or 100&100&1e-5&Not used\\
FA&20 or 100&100&5e-5&Not used\\
DFA&20 or 100&100&5e-5&Not used\\
DDTP&20 or 100&100&3e-4, 9e-4, 1e-4, resp. for each layer&Not used\\
PC&20 or 100&100&1e-5&Not used\\
BP&20 or 100&100&1e-5&Step LR (step size 1)\\
\hline
\end{tabular}
\end{table}
\vspace{-0.1cm}
\begin{table}[h]
\centering
\small
\begin{tabular}{p{1cm}p{1.2cm}p{0.5cm}p{1.5cm}p{1.9cm}}
\hline
\multicolumn{5}{c}{\centering{Hyper-parameters for NLP dataset}} \\ \hline
Algorithm & Epochs & Batch-size & Initial LR & LR Schedule\\
\hline
\multicolumn{5}{c}{\centering{Using 20\% training set}} \\ \hline
HB & 20 or 100 & 100 & 1e-6 & Not used\\
FA & 20 or 100 & 20 & 5e-4 & Not used\\
DFA & 20 or 100 & 20 & 5e-4 & Not used\\
BP & 20 or 100 & 20 & 1e-3 & 0.1x decay at epoch 40 \& 60\\
\hline
\multicolumn{5}{c}{\centering{Using 100\% training set}} \\ \hline
HB & 20 or 100 & 100 & 1e-6 & Not used\\
FA & 20 or 100 & 100 & 5e-4 & Not used\\
DFA & 20 or 100 & 100 & 5e-4 & Not used\\
BP & 20 or 100 & 100 & 1e-3 & 0.1x decay at epoch 40 \& 60\\
\hline
\end{tabular}
\end{table}

\end{document}